\documentclass[letterpaper,10pt,conference]{IEEEtran}
\IEEEoverridecommandlockouts 
\usepackage{amsmath,amssymb,amsfonts,pifont}
\usepackage{hyperref,graphicx}
\usepackage{tabu,float,textcomp}
\usepackage{multirow,dcolumn,hhline}
\usepackage{color,booktabs,relsize}
\usepackage[subrefformat=parens,labelformat=parens,caption=false]{subfig}
\usepackage{adjustbox}
 \usepackage{cancel}
\def\BibTeX{{\rm B\kern-.05em{\sc i\kern-.025em b}\kern-.08em
    T\kern-.1667em\lower.7ex\hbox{E}\kern-.125emX}}
\begin{document}
\title{Real-Time Vibration-Based Bearing Fault Diagnosis Under Time-Varying Speed Conditions}
\author{
    \IEEEauthorblockN{
        Tuomas Jalonen\IEEEauthorrefmark{1},
        Mohammad Al-Sa'd\IEEEauthorrefmark{1}\IEEEauthorrefmark{2},
        Serkan Kiranyaz\IEEEauthorrefmark{3},
        Moncef Gabbouj\IEEEauthorrefmark{1}}
    \IEEEauthorblockA{\IEEEauthorrefmark{1}Department of Computing Sciences, Tampere University, 33720 Tampere, Finland\\Email: \href{mailto:tuomas.jalonen@tuni.fi}{tuomas.jalonen@tuni.fi}; \href{mailto:mohammad.al-sad@tuni.fi}{mohammad.al-sad@tuni.fi}; \href{mailto:moncef.gabbouj@tuni.fi}{moncef.gabbouj@tuni.fi}}
    \IEEEauthorblockA{\IEEEauthorrefmark{2}Department of Physiology, University of Helsinki, 00014 Helsinki, Finland\\Email: \href{mailto:mohammad.al-sad@helsinki.fi}{mohammad.al-sad@helsinki.fi}}
    \IEEEauthorblockA{\IEEEauthorrefmark{3}Department of Electrical Engineering, Qatar University, 2713 Doha, Qatar\\Email: \href{mailto:mkiranyaz@qu.edu.qa}{mkiranyaz@qu.edu.qa}}
    }
\maketitle
\begin{abstract} 
Detection of rolling-element bearing faults is crucial for implementing proactive maintenance strategies and for minimizing the economic and operational consequences of unexpected failures. However, many existing techniques are developed and tested under strictly controlled conditions, limiting their adaptability to the diverse and dynamic settings encountered in practical applications.
This paper presents an efficient real-time convolutional neural network (CNN) for diagnosing multiple bearing faults under various noise levels and time-varying rotational speeds. Additionally, we propose a novel Fisher-based spectral separability analysis (SSA) method to elucidate the effectiveness of the designed CNN model. We conducted experiments on both healthy bearings and bearings afflicted with inner race, outer race, and roller ball faults. 
The experimental results show the superiority of our model over the current state-of-the-art approach in three folds: it achieves substantial accuracy gains of up to 15.8\%, it is robust to noise with high performance across various signal-to-noise ratios, and it runs in real-time with processing durations five times less than acquisition. Additionally, by using the proposed SSA technique, we offer insights into the model's performance and underscore its effectiveness in tackling real-world challenges.
\end{abstract}
\section{Introduction}
\begin{figure}[!t]
\centerline{\includegraphics[width=0.35\textwidth]{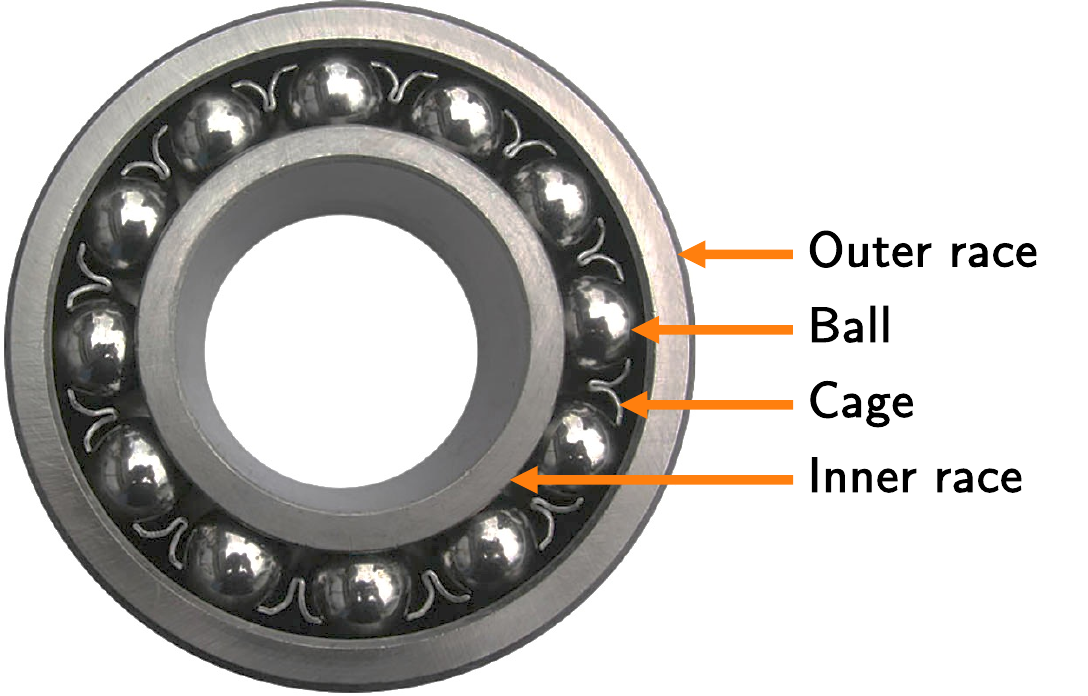}}
    \caption{A rolling element bearing example. The main parts outer race, inner race, balls, and cage are illustrated. Redrawn from \url{https://upload.wikimedia.org/wikipedia/commons/f/f6/HARP_bearing.jpg}.}
    \label{fig:bearing}
\end{figure}

Rolling-element bearings are fundamental components in various industrial machinery and mechanical systems, such as pumps, compressors, and motors, to name a few \cite{alshorman2020review}; see Fig. \ref{fig:bearing}. However, over time, bearings are susceptible to wear, fatigue, and faults that can lead to unexpected machinery failures, production downtime, and costly maintenance \cite{rai2016review}. Hence, the precise diagnosis of bearing faults is of utmost importance \cite{rai2016review}. Fault diagnosis of rolling-element bearings has been a focal point of interest among researchers over several decades \cite{Gunerkar2019}. Data pertinent to this diagnosis can be acquired from various sources, including vibration \cite{Gunerkar2019}, acoustic emission \cite{aasi2022experimental}, oil debris \cite{Zhao2023}, electrostatic measurements \cite{li2020specially}, motor current \cite{9558835}, and visual images \cite{rai2016review}. Nevertheless, studies have shown that vibration is one of the most reliable and cost-effective techniques, and allows capturing subtle changes in dynamics that are associated with early-stage defects \cite{rai2016review,9558835}.

The integration of advanced machine learning technologies with vibration analysis has ushered in a new era for bearing fault diagnosis \cite{alshorman2020review}. Specifically, leveraging Convolutional Neural Networks (CNN) was found to be a promising approach for detecting and identifying bearing defects at their early stages; thereby, enabling proactive maintenance strategies and ultimately reducing the economic and operational impact of unexpected failures \cite{7501527,Eren2019,8584489}. Unfortunately, current solutions are usually designed to work in a very restricted setting and do not integrate the varying conditions and operations found in a real-life scenario, e.g., heavy noise and time-varying rotational speeds.
Therefore, we propose a lightweight real-time CNN network to diagnose bearing faults under various noise levels and time-varying rotational speeds\footnote{The implementation is available at \url{https://github.com/TuomasJalonen/motor-fault-icit2024}.}. In addition, we design a novel Fisher-based separability analysis technique to rationalize the efficacy of the proposed model.
\\
The key contributions of this paper are:
\begin{itemize}
    \item A simple yet effective CNN architecture that exhibits robustness in noisy and variable speed conditions.
    \item A precise real-time diagnostic tool for various bearing faults that surpasses the current state-of-the-art technique \cite{Ni2023} with a great margin; up to 15.8\% gain in accuracy.
    \item A Fisher-based spectral separability analysis technique that surrogates the model's performance and justifies its class-specific efficacy and shortcomings.
\end{itemize}

Conventional signal processing techniques involve manual extraction of features for diagnosing bearing defects \cite{rai2016review}, including time-domain acoustic features \cite{aasi2022experimental}, spectral amplitude modulation and kurtosis \cite{moshrefzadeh2021condition}, spectral segmentation \cite{zhang2021novel}, variational mode decomposition \cite{hu2020rotating,zhou2022rolling}, multi-scale indicators combined with random forest \cite{hu2020rotating} and support vector machines \cite{zhou2022rolling}.
While hand-crafted features are effective in stable environments, they face challenges in noisy conditions, prompting the adoption of automatic feature extraction techniques, notably CNNs \cite{jalonen2023realtime}. For instance, a 2D discrete Fourier transform was performed on the vibration signals in \cite{youcef2020rolling} to create spectral images, which were used for CNN-based classification. Additionally, deep belief networks \cite{shao2015rolling,niu2021optimized}, physics-informed CNN models \cite{sadoughi2019physics,shen2021physics}, and attention mechanisms \cite{plakias2020fault} have all been explored for bearing fault diagnosis. Furthermore, Transformers have demonstrated remarkable fault detection performance but were found computationally intensive and less suitable for smaller datasets \cite{jin2022time}.

Unfortunately, despite the wide array of methods discussed above, they do not consider the case of time-varying rotational speeds i.e., non-stationary signal analysis \cite{9456035}, and tend to train/test only on fixed working conditions. To remedy this limitation, a Transformer with a self-attention module \cite{chi2022bearing} and recurrent neural networks \cite{an2020novel} were suggested to detect rolling-element bearing faults under time-varying speeds. Nevertheless, a new time-varying speed dataset was recently developed at the Korean Advanced Institute of Science and Technology \cite{KAIST_data} and the current state-of-the-art solution for it is the physics-informed residual network (PIResNet) \cite{Ni2023}. Another solution was developed in \cite{anbalagan2023foundational}, but the fault diagnosis problem was simplified into a binary classification to detect healthy and non-healthy bearings; hence, it does not qualify as a bearing diagnostic tool. Consequently, comparative evaluations are performed against the state-of-the-art PIResNet approach \cite{Ni2023}.
\section{Methodology} \label{sec:methodology}
A rolling-element bearing is comprised of rollers that serve to uphold the spatial separation between two bearing races \cite{Kumar2018}; see Fig. \ref{fig:bearing}. The primary function of a ball bearing is to mitigate rotational friction while providing support for both radial and axial loads \cite{Kumar2018}. This is accomplished by employing a minimum of two races to encase the balls and facilitate the transmission of loads \cite{Kumar2018}. Faults in bearings represent the predominant category of anomalies occurring in induction machines; hence, continuous monitoring is essential \cite{DeekshitKompella2018}. Fortunately, vibration in machines is mainly generated due to defective bearings; hence, analyzing its morphology allows us to diagnose the bearing's health state \cite{youcef2020rolling,Malla2019}.
\subsection{Dataset description}
We utilized an open-access dataset collected from a bearing test rig at the Korean Advanced Institute of Science and Technology (KAIST) \cite{KAIST_data}. The test rig consists of a three-phase induction motor, torque meter, gearbox, bearing housing A, bearing housing B, rotors, and a hysteresis brake \cite{KAIST_data}. Four ceramic shear ICP-based accelerometers were mounted on the x- and y-directions of the two bearing housings \cite{KAIST_data}. The motor was operated with time-varying speeds (680-2460 rpm) and the bearing in housing B was seeded with outer race, inner race, and bearing ball faults. These experiments yielded four classes named \emph{Normal}, \emph{Outer}, \emph{Inner}, and \emph{Ball}, describing the bearing's normal operation and three common faults.
The vibration signals under time-varying speed conditions were acquired by the accelerometers installed on housing B. The sampling frequency was set to 25.6 kHz and 2,100 seconds of vibration data were collected for each of the four classes.
\subsection{Signal preprocessing}
\begin{figure}[!t]
\centerline{\includegraphics[width=0.375\textwidth]{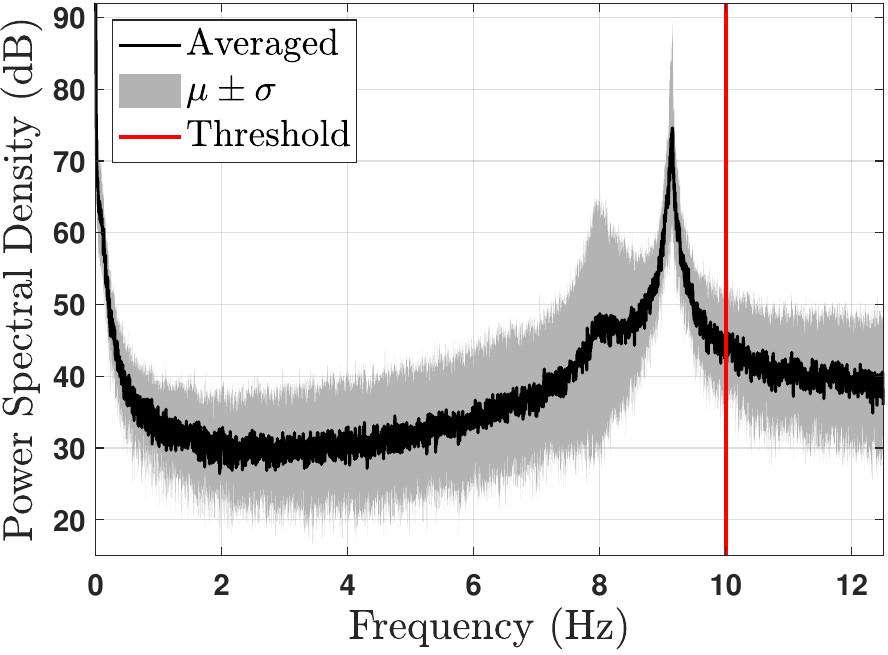}}
   \caption{Frequency analysis of the motor's rotational speed. The power spectral density was estimated by the nonuniform Fourier transform (NUFT) because of the speed's inconsistent acquisition rate. Frequency scaling for the NUFT was performed using a sampling frequency of 12.5 Hz; the reciprocal of the shortest acquisition time. The results were averaged across the different classes and dataset files and plotted with their 68.3\% confidence intervals.}
   \label{fig:segment_length}
\end{figure}

The vibration signals from the two sensors were first downsampled to 20 kHz using a finite impulse response anti-aliasing lowpass filter with delay compensation ($f_s=20000$). 
This operation yields a discrete multi-sensor signal $\boldsymbol{s}=[s_1(n), s_2(n)]^T$ of size $2\times N$ where $s_q(n) = \left[s_q(1), s_q(2), \cdots, s_q(N)\right]$ is the signal acquired by sensor $q$ with $N=42\times10^6$ time samples \cite{Boashash2018}.
After that, we formed a noisy multi-sensor vibration signal $\boldsymbol{x}$ as:
\begin{equation}
    \boldsymbol{x} = \boldsymbol{s} + \alpha\,\boldsymbol{\eta}\,,
\end{equation}
where $\boldsymbol{\eta} =[\eta_1(n), \eta_2(n)]^T$ holds independent and identically distributed real-valued noise, $\eta_q(n) \sim \mathcal{N}(0,\,1)$ is a zero-mean white Gaussian noise with unit variance, and $\alpha = [\alpha_1, \alpha_2]^T$ are factors that control the amount of noise to be added to $\boldsymbol{s}$ given a pre-defined signal-to-noise ratio (SNR), i.e.:
\begin{equation}
    \alpha_q = \sqrt{\dfrac{10^{-\text{SNR}/10}}{N}\sum_{n=1}^{N} s_q^2(n)}\,.
\end{equation}
In this study, we conducted experiments utilizing both clean vibration signals and those subjected to various degrees of noise-induced degradation. Specifically, we corrupted the vibration waveforms with noise at the following SNR levels: 20 dB, 15 dB, 10 dB, 5 dB, 0 dB, and -5 dB, to span the range of mild to moderate to severe noise conditions.\\
Furthermore, we divided the temporal course of the signals in $\boldsymbol{x}$ into smaller segments of length $L$ with no overlap to prevent any inadvertent data leakage when training and/or testing the CNN model.
The segmented signal $\boldsymbol{v}$ is expressed as a 3-way tensor of size $2\times P \times L$, i.e.:
\begin{equation}
    \boldsymbol{v} = 
    \left[\begin{array}{cccc}
    v_{1,1}(\ell)  & v_{1,2}(\ell) & \cdots & v_{1,P}(\ell)\\
    v_{2,1}(\ell)  & v_{2,2}(\ell) & \cdots & v_{2,P}(\ell)
    \end{array}\right]\,,
\end{equation}
\begin{equation}
\footnotesize
    v_{q,p}(\ell) = \left[x_q\left(L(p-1)+1\right), x_q\left(L(p-1)+2\right), \cdots, x_q\left(pL\right)\right]\,,
\end{equation}
where $P=N/L$ is the total number of segments with length $L$, $p\in[1,P]$, and $\ell\in[1,L]$.
The segment length $L$ is a paramount parameter that needs careful selection. On the one hand, it needs to be small enough to generate a sufficient number of samples for training and to minimize additional complexities emerging from the motor's varying speed. On the other hand, it should be wide enough to include the motor's vibration complexities and to learn class-specific patterns.
To determine an adequate segment length, we conducted a frequency analysis on the motor's speed to identify its maximum rate of change; see Fig. \ref{fig:segment_length}. The analysis showed that the motor's rotational speed varies at two specific frequencies; 8 Hz and 9.15 Hz. Therefore, we decided 10 Hz, or 0.1 seconds, is a suitable maximum rate of range which includes 99.6\% of the total power spectrum; almost all rapid changes in rotational speed. The selected threshold corresponds to a segment length of 2,000 samples ($L=2000$) which yields a total of 21,000 segments ($P=21000$) which is ample for training and/or testing deep learning models.
\\
Finally, the generated segments in $\boldsymbol{v}$ exhibit variations in scale. Therefore, standardization is imperative for rigorous and precise processing. In this study, we calculated the standardized, or z-score normalized, tensor $\boldsymbol{z}$ as follows:
\begin{equation}
    \boldsymbol{z} = 
    \left[\begin{array}{cccc}
    z_{1,1}(\ell)  & z_{1,2}(\ell) & \cdots & z_{1,P}(\ell)\\
    z_{2,1}(\ell)  & z_{2,2}(\ell) & \cdots & z_{2,P}(\ell)
    \end{array}\right]\,,
\end{equation}
\begin{equation}
    z_{q,p}(\ell) = \left(v_{q,p}(\ell)-\mu_p\right)/\sigma_p\,,
\end{equation}
where $\mu_p$ and $\sigma_p$ are the mean and standard deviation of the $p$-th segment from both sensors (flattened), respectively.
\subsection{The deep learning model}
We designed a lightweight CNN model to allow the utility of inexpensive edge computing units for deployment. The network accepts the standardized vibration signals $\boldsymbol{z}$ coming from both accelerometers (mounted on the x- and y-directions) of housing B, i.e., the network's input size is $2000\times2$ (input signal is transposed). Afterward, the input signals are passed through five convolutional blocks each containing a convolutional layer for feature extraction with a ReLU activation function to account for non-linear effects, and a max-pooling layer of size $2\times1$ to summarize the feature maps. The convolution layer in the first block has a kernel size of $5\times1$ while the rest have $3\times1$ kernels. The extracted features (coming from the last convolutional block) are flattened and passed through a dense layer with $128$ nodes and a ReLU activation function. Finally, they are fed to a dropout layer with a rate of $0.4$ and a classification layer with $4$ nodes and a Softmax activation function to transform the network's raw outputs into a vector of probabilities.
In total, the model has 558,660 trainable parameters.

Training and testing the CNN model was conducted in a 5-fold cross-validation to accurately estimate its ability to generalize to an independent dataset. First, we divided the 84,000 samples (four classes each having 21,000 segments) into five equally sized shuffled splits and maintained their class balance (stratified folds). Out of the five splits, a single split was kept for testing, while the remaining four splits were used for training. We repeated this process five times, with each of the five splits used exactly once for testing. Specifically, each model was trained with 67,200 samples and tested with 16,800 samples. 
Training underwent for 150 epochs, to ensure convergence, and Adam optimizer was employed to minimize the categorical cross-entropy loss \cite{kingma2014adam}, i.e.:
\begin{equation}
    \mathcal{L}= -y \log(\hat{y}) - (1-y) \log(1 - \hat{y})\,,
\end{equation}
where $\mathcal{L}$ denotes the loss, $y$ and $\hat{y}$ are the true and predicted class labels, respectively.
We set the batch size to 100, and the learning rate to $10^{-5}$ for epochs 1-101 and $10^{-6}$ for epochs 102-150. Finally, we tested the CNN model using the weights from the last epoch (no early stopping), and the same structure and training settings were used for all noise levels.
\subsection{The spectral separability analysis}
The rationale for the efficacy of a deep learning model or elucidation of the underlying factors contributing to its shortcomings is imperative for its seamless integration and advancement. In this study, we propose a Fisher-based spectral separability analysis (SSA) to quantify the presence of discriminative features in the vibration signals. The SSA facilitates differentiating the distinct bearing health states and serves as a surrogate indicator for the model's performance. For instance, if there is a spectral distinction between two classes, the model should be capable of detecting it, whereas if such disparity is absent, the model's expected/unexpected limitations/efficacy becomes apparent.
The SSA starts by calculating the smoothed power spectral density (PSD) of the vibration signal $z_{q,p}^{(i)}(\ell)$ belonging to class $i$ by:
\begin{equation}
    \small
    \hat{z}_{q,p}^{(i)}(k) = 10\log_{10}\left( \dfrac{1}{L-1}\left|\underset{\ell \rightarrow k}{\mathcal{F}}\big\{z_{q,p}^{(i)}(\ell)\big\}\right|^2\underset{k}{*}\,\omega_h(k) \right)\,,
\end{equation}
where $\hat{z}_{q,p}^{(i)}(k)$ is the PSD of $z_{q,p}^{(i)}(\ell)$, $\underset{\ell \rightarrow k}{\mathcal{F}}$ is the Fourier transform operator from the discrete time $\ell$ to discrete frequency $k$, $\underset{k}{*}$ denotes linear convolution in $k$, and $\omega_h(k)$ is a smoothing window of length $h$ such that $\omega_h(k)\neq0$ for $|k|\leq h/2$, $\omega_h(k)=\omega_h(-k)$, and $\omega_h(k_1)>\omega_h(k_2)$ for $|k_1|<|k_2|$.
After that, the Fisher criterion (FC) for measuring the separability between classes' $i$ and $j$ power spectra is expressed as:
\begin{equation}
    f_{q}^{(i,j)}(k) = \dfrac{\left|\mu_{q,i}(k)-\mu_{q,j}(k)\right|^2}{\sigma_{q,i}^2(k)+\sigma_{q,j}^2(k)}\,,
\end{equation}
where $\mu_{q,i}(k)$ and $\sigma_{q,i}^2(k)$ are the mean and variance of $\boldsymbol{\hat{z}}_{q,p}^{(i)}(k)$, respectively, i.e.:
\begin{equation}
    \mu_{q,i}(k) = \dfrac{1}{P}\sum_{p=1}^P\hat{z}_{q,p}^{(i)}(k)\,.
\end{equation}
\begin{equation}
    \sigma_{q,i}^2(k) = \dfrac{1}{P}\sum_{p=1}^P \left(\hat{z}_{q,p}^{(i)}(k)-\mu_{q,i}(k)\right)^2\,.
\end{equation}
Finally, the FCs are thresholded and smoothed to retain highly separable continuous frequency bands, i.e.:
\begin{equation}
    \zeta_{q}^{(i,j)}(k) = 
    \left(
        \begin{cases}
        f_{q}^{(i,j)}(k) &: f_{q}^{(i,j)}(k) > \epsilon\\
        \quad\,\,\,0 &: \,\,\text{otherwise}
    \end{cases}
    \right)\underset{k}{*}\,\omega_h(k)\,,
\end{equation}
where we set the threshold $\epsilon$ to 2 to allow a maximum of 31.73\% overlap between the classes' probability distributions.
\subsection{Performance evaluation and analysis}
The performance of the CNN model was evaluated by employing a one-versus-all approach, thereby transforming the initial multi-class problem into a set of binary classification tasks.
The assessment of each binary task was quantified by accuracy, precision, recall, and F1-score, and then these metrics were averaged across all binary tasks (macro-averaged) to assess the multi-class performance, i.e.:
\begin{equation}
    \text{Accuracy} = \dfrac{1}{I}\sum_{i=1}^I \left(\dfrac{TP_i+TN_i}{TP_i+TN_i+FP_i+FN_i}\right)\,,
\end{equation}
\begin{equation}
    \text{Precision} = \dfrac{1}{I}\sum_{i=1}^I \left(\dfrac{TP_i}{TP_i+FP_i}\right)\,,
\end{equation}
\begin{equation} \label{eq:recall}
    \text{Recall} = \dfrac{1}{I}\sum_{i=1}^I \left(\dfrac{TP_i}{TP_i+FN_i}\right)\,,
\end{equation}
\begin{equation}
    \text{F1-Score} = \dfrac{1}{I}\sum_{i=1}^I \left(\dfrac{2\,TP_i}{2\,TP_i+FN_i+FP_i}\right)\,,
\end{equation}
where $I=4$ is the total number of classes, $TP_i$, $TN_i$, $FP_i$, and $FN_i$ are true positives, true negatives, false positives, and false negatives, respectively, when detecting class $i$ and rejecting all other classes.
\\
Moreover, we analyzed the predictive capacity of the CNN model by t-distributed stochastic neighbor embedding (t-SNE); a dimensionality reduction method that clusters similar high dimensional samples and departs dissimilar ones in 2- or 3-dimensional space \cite{van2008visualizing}. This technique helps in characterizing the model's predictive power when supplied with new vibration signals. In other words, given large enough training samples, if the t-SNE shows clear inter-class separability, one may infer the adequacy of the model for unseen samples.
\\
Finally, the complexity of the CNN model was assessed by its inference time. We conducted Monte-Carlo simulations where we fed the model with 1000 test samples, predicted their class, and repeated the process 10 times for validation. We used a MacBook Pro with ARM-based M1 Pro chip, integrated 16-core GPU, 16-core neural engine, and 16 GB of RAM.
\section{Results and discussion} \label{sec:results}
\subsection{Fault diagnosis performance}
\begin{figure}[!t]
\centerline{\includegraphics[width=0.4\textwidth]{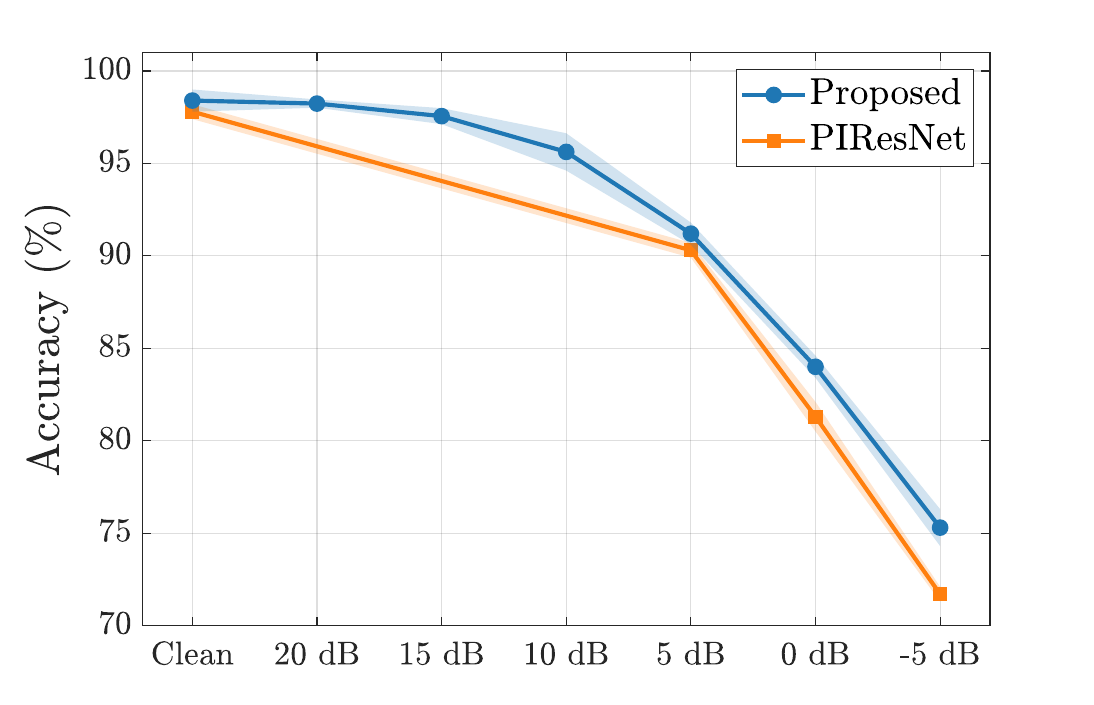}}
   \caption{Comparing the testing macro-averaged accuracy of the CNN network to the current state-of-the-art technique, the PIResNet \cite{Ni2023}, using both clean and noisy vibration signals. The accuracy curves are plotted at varying SNR levels along with their 95\% confidence intervals ($\mu\pm2\sigma$).}
   \label{fig:accuracy}
\end{figure}

\begin{figure*}[!t]
   \centering
   \subfloat[Clean data (no added noise). \label{fig:cm_1}]{\includegraphics[width=.24\textwidth]{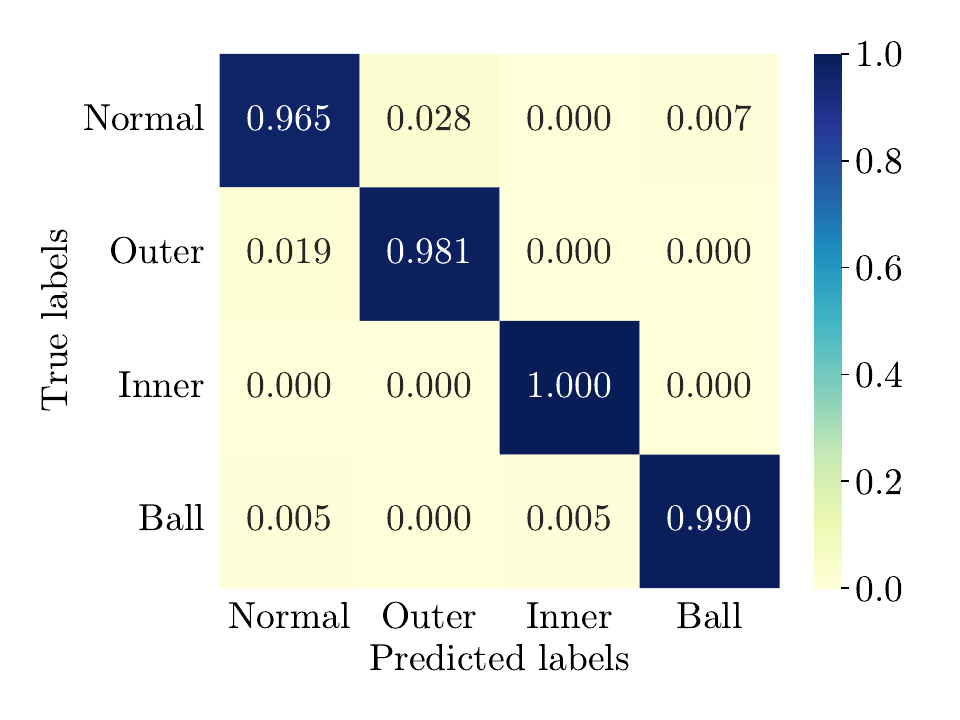}}\,
   \subfloat[SNR = 5 dB. \label{fig:cm_2}]{\includegraphics[width=.24\textwidth]{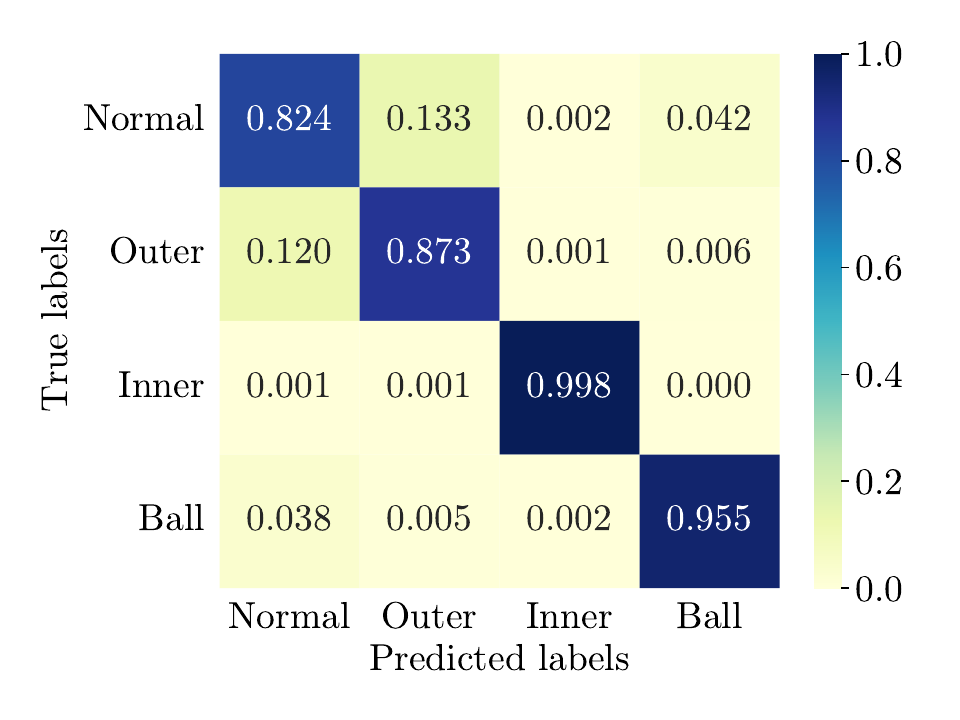}}\,
   \subfloat[SNR = 0 dB. \label{fig:cm_3}]{\includegraphics[width=.24\textwidth]{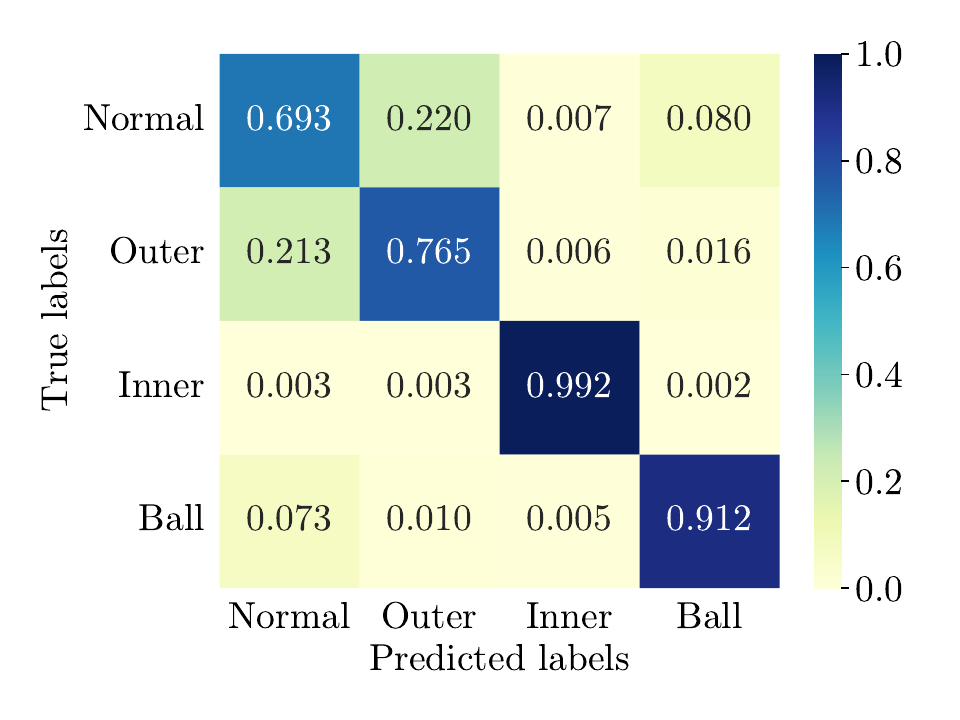}}\,
   \subfloat[SNR = -5 dB. \label{fig:cm_4}]{\includegraphics[width=.24\textwidth]{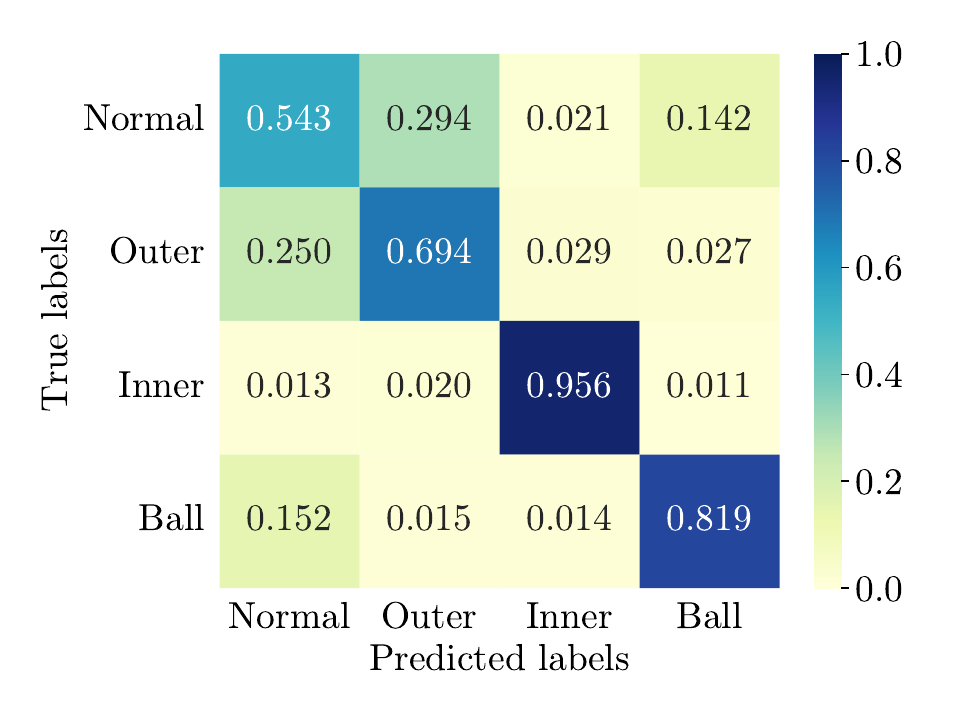}}
   \caption{The testing confusion matrices averaged over the five data splits. The results depict the performance when using (a) clean, and (b)-(d) noisy vibration signals at the following SNR levels: (b) 5 dB, (c) 0 dB, and (d) -5 dB.}
   \label{fig:cms}
\end{figure*}

Fig. \ref{fig:accuracy} demonstrates the testing accuracy of the designed CNN network for classifying the four health states of the bearing across the different SNR levels. Additionally, it illustrates a comparison between our proposed model and the current state-of-the-art approach, PIResNet \cite{Ni2023}. The results indicate that, whether in clean or severe noisy signal conditions, the proposed approach surpasses the performance of the PIResNet (up to a 3.6\% point gain in average accuracy). The extent of this performance enhancement inversely correlates with the SNR level, meaning that the superiority of the proposed approach becomes more pronounced as the level of noise in the vibration signal increases.
Table \ref{table:tab_perf_results} confirms this observation by extending the comparison to include the complete set of testing metrics for both models; accuracy, precision, recall, and the F1-score. Specifically, it reveals a persistent outperformance of the proposed CNN in comparison to the PIResNet across all metrics and SNR levels, thus rendering the additional complexity associated with the PIResNet redundant.
\begin{table*}[!t]
\footnotesize
\centering
\caption{The testing performance in terms of macro-averaged accuracy, precision, recall, and F1-score. The metrics are summarized by their 5-fold averaged percentages $\pm$ standard deviations and the best outcomes are highlighted in bold. The precision, recall, and F1-score of PIResNet are calculated from the reported best-performing confusion matrices in \cite{Ni2023}; thus, the results' standard deviations are missing.}
\begin{tabular}{ccccccccc}
\toprule
& \multicolumn{2}{c}{\textbf{Clean}} & \multicolumn{2}{c}{\textbf{SNR = 5 dB}} & \multicolumn{2}{c}{\textbf{SNR = 0 dB}} & \multicolumn{2}{c}{\textbf{SNR = -5 dB}} \\
\cmidrule{2-9}
& \multicolumn{1}{c}{\textbf{Proposed}} & \multicolumn{1}{c}{\textbf{PIResNet \cite{Ni2023}}} & \multicolumn{1}{c}{\textbf{Proposed}} & \multicolumn{1}{c}{\textbf{PIResNet \cite{Ni2023}}} & \multicolumn{1}{c}{\textbf{Proposed}} & \multicolumn{1}{c}{\textbf{PIResNet \cite{Ni2023}}} & \multicolumn{1}{c}{\textbf{Proposed}} & \multicolumn{1}{c}{\textbf{PIResNet \cite{Ni2023}}}
\\\midrule
\textbf{Accuracy} & $\mathbf{98.4 \pm 0.3}$ & $97.8 \pm 0.2$ & $\mathbf{91.2 \pm 0.3}$ & $90.3 \pm 0.2$ & $\mathbf{84.0 \pm 0.3}$ & $81.3 \pm 0.4$ & $\mathbf{75.3 \pm 0.5}$ & $71.7 \pm 0.2$
\\\midrule
\textbf{Precision} & $\mathbf{98.4 \pm 0.3}$ & $98.0 \,\pm$ N/A & $\mathbf{91.3 \pm 0.4}$ & $90.5 \,\pm$ N/A & $\mathbf{84.0 \pm 0.2}$ & $81.6 \,\pm$ N/A & $\mathbf{75.1 \pm 0.7}$ & $72.3 \,\pm$ N/A
\\\midrule
\textbf{Recall} & $\mathbf{98.4 \pm 0.3}$ & $98.0 \,\pm$ N/A & $\mathbf{91.2 \pm 0.4}$ & $90.5 \,\pm$ N/A & $\mathbf{84.0 \pm 0.3}$ & $81.6 \,\pm$ N/A & $\mathbf{75.3 \pm 0.6}$ & $72.0 \,\pm$ N/A
\\\midrule
\textbf{F1-Score} & $\mathbf{98.4 \pm 0.3}$ & $98.0 \,\pm$ N/A & $\mathbf{91.2 \pm 0.4}$ & $90.5 \,\pm$ N/A & $\mathbf{84.0 \pm 0.3}$ & $81.6 \,\pm$ N/A & $\mathbf{75.2 \pm 0.7}$ & $72.2 \,\pm$ N/A
\\\bottomrule
\end{tabular}
\label{table:tab_perf_results}
\end{table*}
\\
Furthermore, Fig. \ref{fig:cms} details the testing performance by presenting the confusion matrices when using clean and noisy vibration signals at three SNR levels: 5 dB, 0 dB, and -5 dB.
The results reveal that:
(1) the \emph{Inner} class is the easiest to predict as it had the least amount of accuracy drop; from 100\% for clean signals to 95.6\% for noisy signals at -5 dB SNR; (2) differentiating the \emph{Normal} and \emph{Outer} states becomes particularly challenging at low SNR, due to the morphological features of these states being predominantly indistinguishable noise; (3) a moderate misclassification between the \emph{Normal} and \emph{Ball} states exists at -5 dB; and (4) by examining the confusion matrices' progression through the SNR levels, one notes a skewness towards the \emph{Inner} and \emph{Ball} states that increases with the amount of noise. This performance imbalance suggests that the temporal morphology of these states is mainly impulsive; hence, they are less affected by additive white Gaussian noise. Finally, when contrasting the results in Fig. \ref{fig:cms} with those reported in \cite{Ni2023}, it becomes apparent that the designed CNN outruns the PIResNet with a significant performance gain of up to 15.8\% points. Nevertheless, our results are averaged over the five cross-validation splits while PIResNet results are drawn from the best-performing confusion matrices. Therefore, the former comparison constitutes a lower limit on the performance gain we can achieve.
\subsection{The t-SNE and complexity analysis}
\begin{figure*}[!t]
   \centering
   \subfloat[Clean data (no added noise). \label{fig:t-sne_1}]{\includegraphics[width=.24\textwidth]{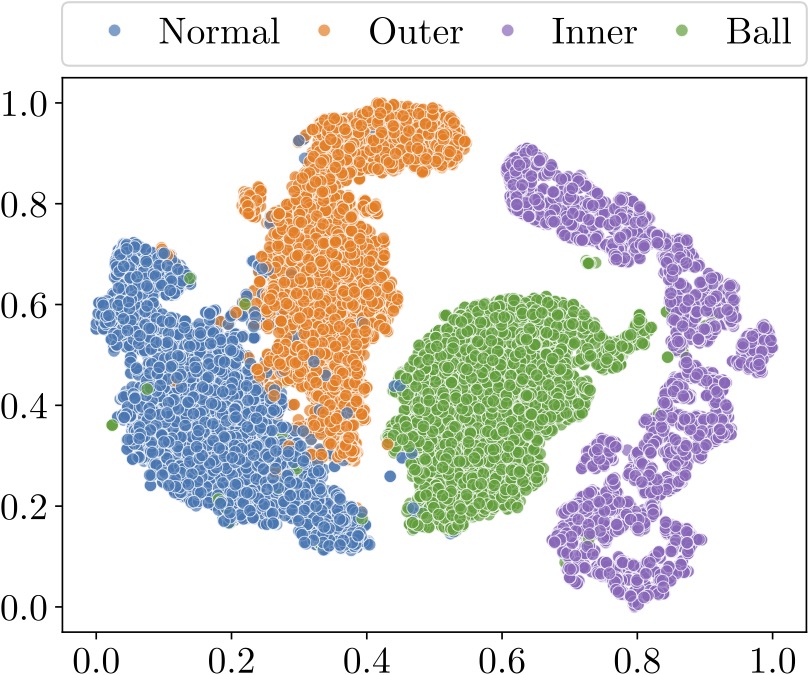}}\,
   \subfloat[SNR = 5 dB. \label{fig:t-sne_2}]{\includegraphics[width=.24\textwidth]{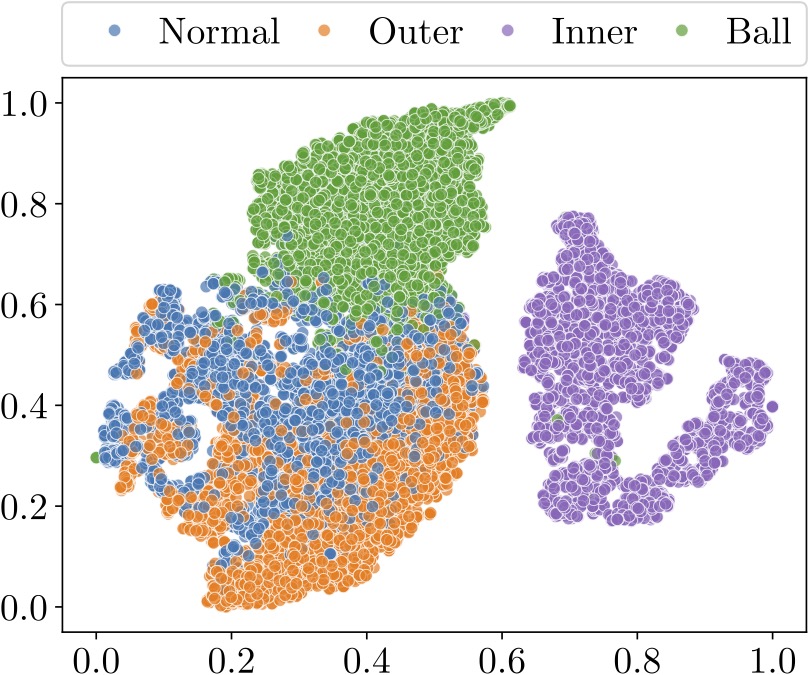}}\,
   \subfloat[SNR = 0 dB. \label{fig:t-sne_3}]{\includegraphics[width=.24\textwidth]{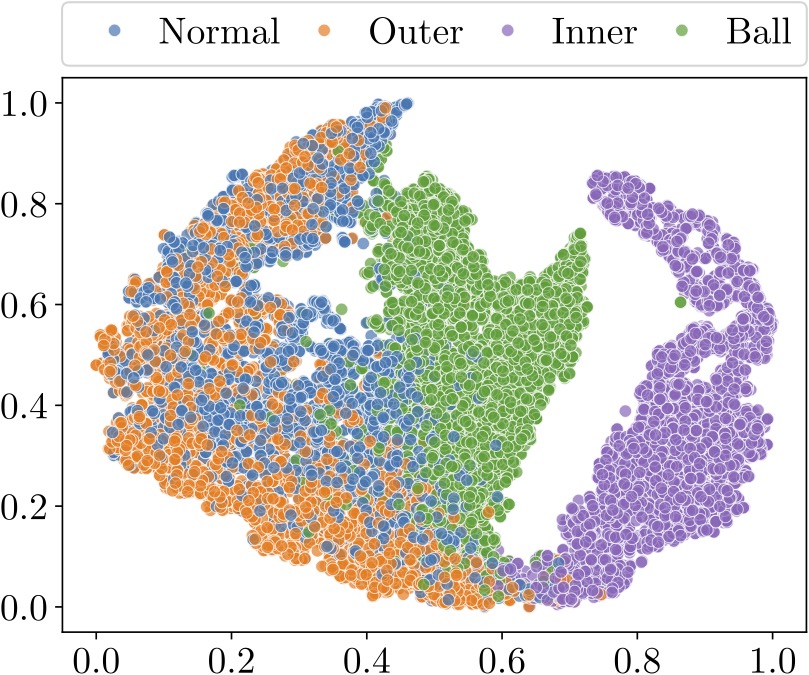}}\,
   \subfloat[SNR = -5 dB. \label{fig:t-sne_4}]{\includegraphics[width=.24\textwidth]{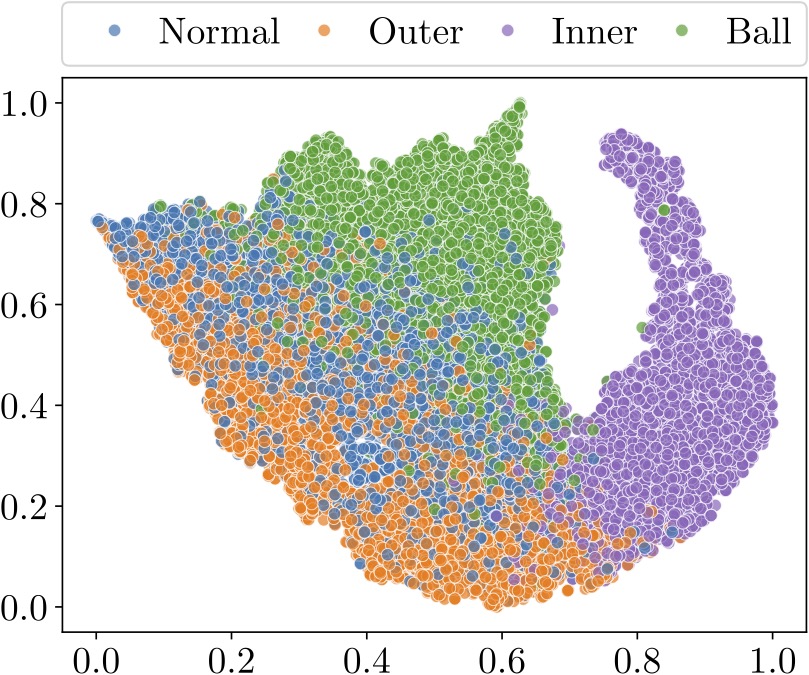}}
   \caption{The model's testing t-SNE from the first data split. The results depict the t-SNE when using (a) clean, and (b)-(d) noisy data at the following SNR levels: (b) 5 dB, (c) 0 dB, and (d) -5 dB. We used the model's last layer features as input to the t-SNE algorithm.}
   \label{fig:t-snes}
\end{figure*}

\begin{figure}[!t]
\centerline{\includegraphics[width=0.4\textwidth]{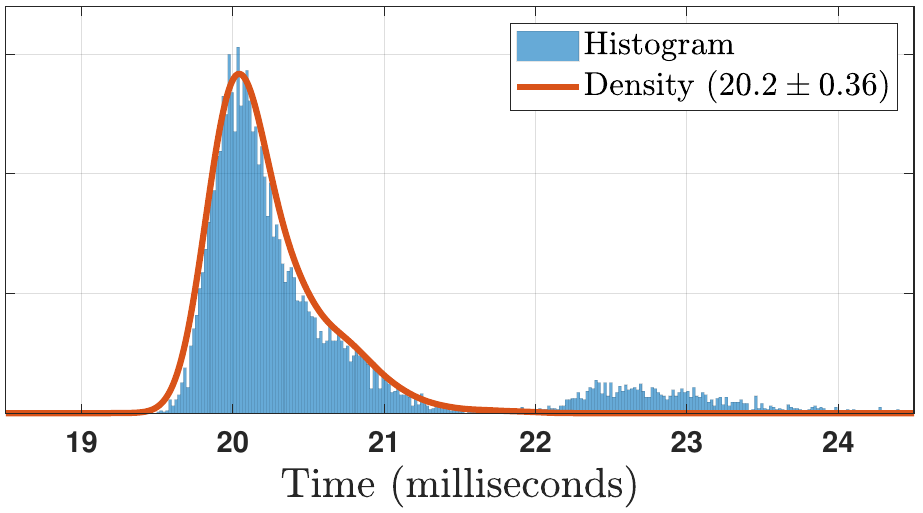}}
   \caption{Statistics of the proposed model's inference time.}
   \label{fig:time}
\end{figure}

Fig. \ref{fig:t-snes} shows the t-SNE results which provide insight into the learned features of the CNN model, where they show the model's ability to learn highly discriminative features when supplied with clean data, and its resilience to noise when distinguishing some of the bearing's health states.
On the one hand, the results in Fig. \ref{fig:t-sne_1} reveal a clear inter-class separation among all health states when using clean vibration signals. On the other hand, Figs. \ref{fig:t-sne_2}-d demonstrate that inter-class separability diminishes in correlation to the level of introduced noise. In particular, we found that the \emph{Inner} class features form the most isolated cluster at all SNR levels, the feature overlap between the \emph{Normal} and \emph{Ball} classes increases with noise, and that features from the states \emph{Normal} and \emph{Outer} intertwine to become indistinguishable from each other when noise is introduced.
The latter behavior is expected as it is a consequence of the noted model's performance imbalance at low SNR levels; see Fig. \ref{fig:cm_4}.
Moreover, the model inference time statistics are demonstrated in Fig. \ref{fig:time}. The results indicate an average inference time of 20.2 ms with a 0.36 standard deviation. This corresponds to a processing time approximately five times less than the duration of data acquisition. In other words, it takes $\approx$ 20 ms to process 100 ms of vibration data coming from the two sensors.
\subsection{The spectral separability analysis}
\begin{figure}[!t]
   \centering
   \subfloat[Clean data (no added noise). \label{fig:f1}]{\includegraphics[width=.45\textwidth]{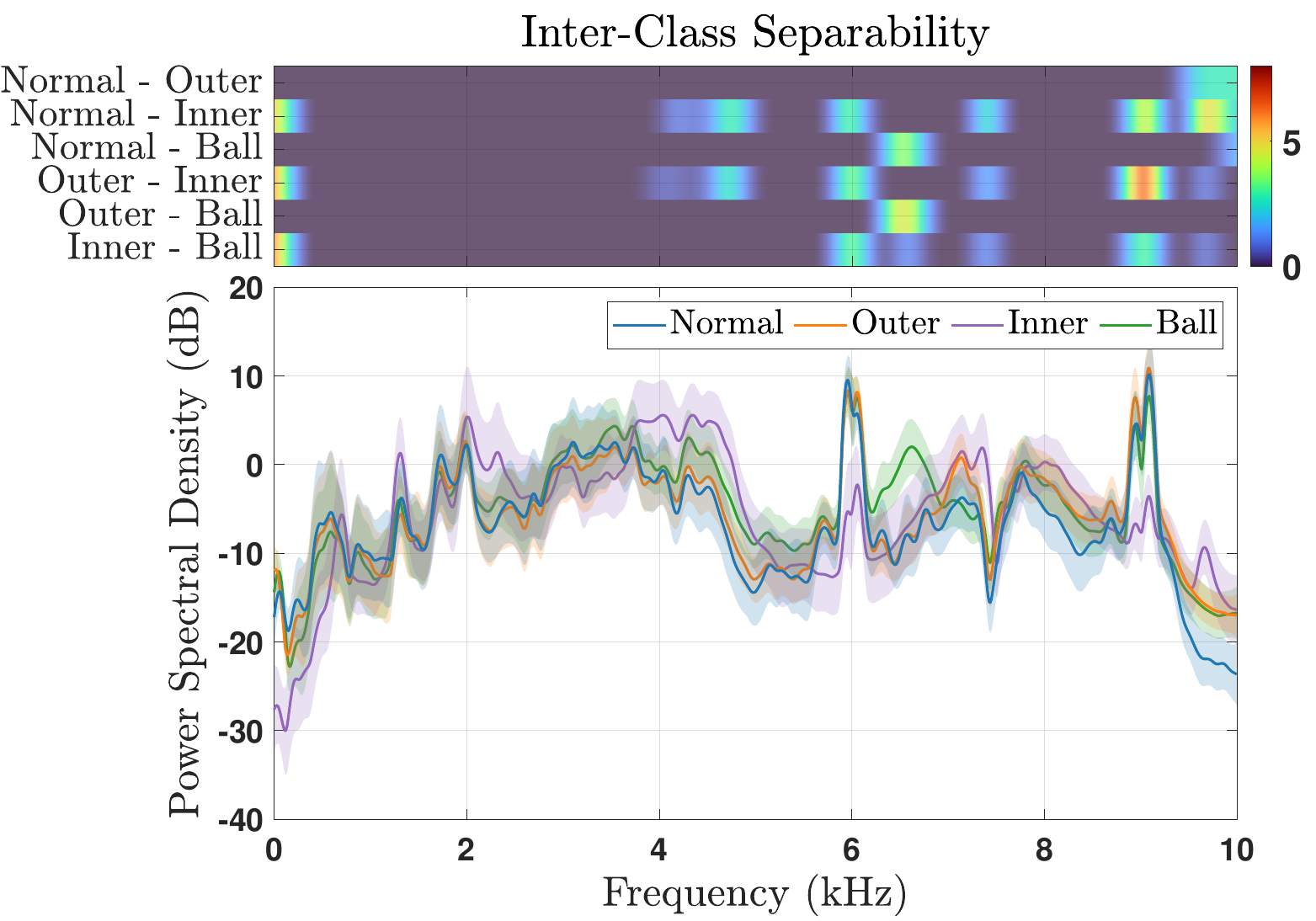}}
   \\
   \subfloat[SNR = 0 dB. \label{fig:f2}]{\includegraphics[width=.45\textwidth]{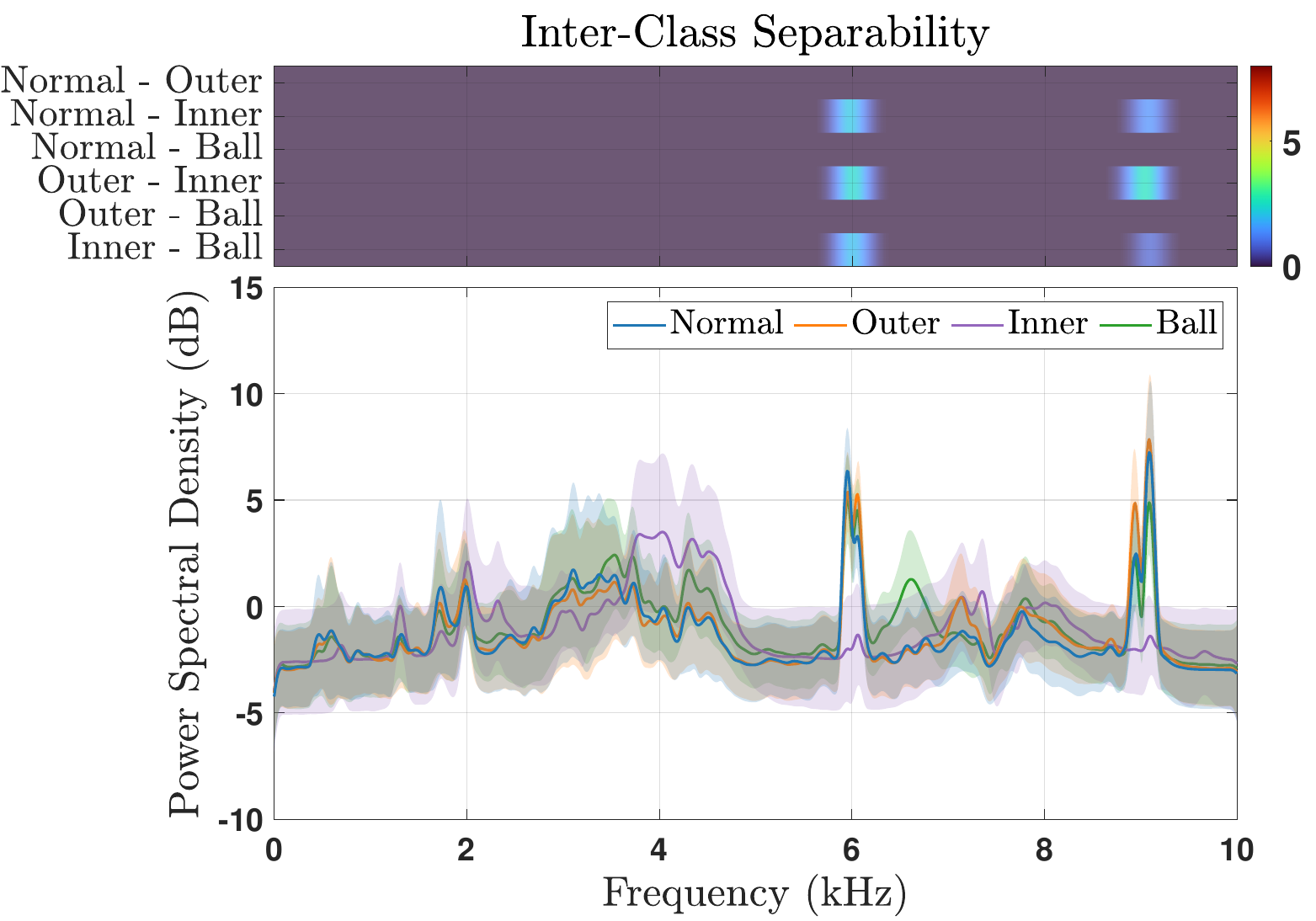}}
   \caption{The SSA of the y-directional vibrations (a) without noise, and (b) with noise at 0 dB SNR. The Fourier transforms were calculated with 1024 samples and smoothed via a Gaussian-weighted moving average filter of length 16 samples. The averaged PSDs are shown for each class along with their 68.3\% confidence intervals ($\mu\pm\sigma$). Inter-class separability was estimated for each pair of groups in a one-versus-one fashion. The SSA estimates were smoothed by a Gaussian-weighted moving average filter of length 64 samples to highlight wide rather than narrow separable bands.}
   \label{fig:frequency_analysis}
\end{figure}

Fig. \ref{fig:frequency_analysis} illustrates the outcomes of the SSA conducted on both the uncontaminated and noisy (SNR = 0 dB) vibration signals, offering a method for explaining the CNN model's efficacy and pitfalls. In the context of clean data, as displayed in Fig. \ref{fig:f1}, the findings reveal the following marked differences in spectral patterns:
(1) the spectral characteristics of the \emph{Inner} class exhibit pronounced dissimilarity when compared to all other classes, notably within the frequency bands $0-0.3$ kHz, $3.8-5$ kHz, $5.8-6.2$ kHz, $7.3-7.5$ kHz, $8.9-9.2$ kHz, and $9.6-9.8$ kHz; (2) the \emph{Ball} class displays moderate spectral distinctions relative to the \emph{Normal} and \emph{Outer} classes, as evident around $6.4-6.8$ kHz; and (3) the \emph{Normal} and \emph{Outer} classes exhibit slight difference around $9.4-10$ kHz. This implies that the high performance demonstrated in Fig. \ref{fig:cm_1} is a consequence of the model's efficacy in acquiring the temporal and spectral knowledge of the non-stationary vibration signals.
On the other hand, in the case of noisy data, as shown in Fig. \ref{fig:f2}, the SSA results show that:
(1) no significant inter-class separability exists apart from one group, the \emph{Inner} class, as evident within $5.8-6.2$ kHz and $8.9-9.2$ kHz; and (2) the separability of the \emph{Inner} class became less significant when compared to the clean case.
These observations explain the model's shortcomings at low SNR levels and its performance imbalance in three folds. First, the \emph{Normal} and \emph{Outer} classes had little discriminative information in the frequency domain; hence, the model relied mainly on their temporal morphology. However, this dependence becomes unreliable when noise is introduced; therefore, features for these two classes become intertwined and dominate the errors as shown in Fig. \ref{fig:cm_3} and Fig. \ref{fig:t-sne_3}. Moreover, the \emph{Inner} state had ample spectral disparity or unique features, which allowed high performance even at low SNR levels. Finally, the \emph{Ball} class had moderate discriminative features in the frequency domain; therefore, the model had to rely on both its temporal and spectral morphology, resulting in a reasonable performance drop at low SNR. Nevertheless, this suggests that vibration signals should be represented in the joint time-frequency domain to exploit their degrees of freedom fully and to design highly robust bearing fault detectors.
\section{Conclusions}
Diagnosing faults in bearings at their early stages allows proactive maintenance strategies and reduces the economic and operational impact of unexpected failures. Unfortunately, current fault diagnosis techniques are often tested in restricted settings and do not integrate the variable conditions and operations found in a real-life scenario.
In this paper, we proposed a vibration-based efficient real-time convolutional neural network (CNN) to diagnose different bearing faults under various noise levels and time-varying rotational speeds. Besides, we designed a novel Fisher-based spectral separability analysis (SSA) technique to rationalize the efficacy of the proposed model.
The experimental results showed that the proposed method outperforms PIResNet, the current state-of-the-art technique, across multiple signal-to-noise ratios with accuracy gains reaching up to 15.8\% while running in real-time.
Moreover, through the SSA outcomes, we explained the model's behavior under heavy noise and indicated the need for representing the vibration signals in the joint time-frequency domain to design robust bearing diagnostic tools. 
The present work has some limitations; it is confined to fixed noise levels and to diagnosing isolated faults. Hence, one may include non-stationary noise and multiple faults occurring simultaneously. 
Other than that, our future research directions include testing the proposed CNN on different datasets, designing a time-frequency-based bearing fault diagnosis system, and incorporating advanced techniques such as operational neural networks \cite{kiranyaz2020operational} with the recent neuron models \cite{kiranyaz2021self}.
\section*{Acknowledgment}
This work was funded by NSF CBL and Business Finland AMALIA project.
\bibliography{Sections/references}
\end{document}